# AI-Driven Road Maintenance Inspection v2: Reducing Data Dependency & Quantifying Road Damage

Haris Iqbal[1]   Hemang Chawla   Arnav Varma   Terence Brouns
Ahmed Badar   Elahe Arani   Bahram Zonooz

Advanced Research Lab, NavInfo Europe, The Netherlands

## ABSTRACT

Road infrastructure maintenance inspection is typically a labor-intensive and critical task to ensure the safety of all road users. Existing state-of-the-art techniques in Artificial Intelligence (AI) for object detection and segmentation help automate a huge chunk of this task given adequate annotated data. However, annotating videos from scratch is cost-prohibitive. For instance, it can take an annotator several days to annotate a 5-minute video recorded at 30 FPS. Hence, we propose an automated labelling pipeline by leveraging techniques like few-shot learning and out-of-distribution detection to generate labels for road damage detection. In addition, our pipeline includes a risk factor assessment for each damage by instance quantification to prioritize locations for repairs which can lead to optimal deployment of road maintenance machinery. We show that the AI models trained with these techniques can not only generalize better to unseen real-world data with reduced requirement for human annotation but also provide an estimate of maintenance urgency, thereby leading to safer roads.

## INTRODUCTION

Road infrastructure maintenance is a critical task to ensure safety of all road users. Traditionally, the damage inspection is done manually, introducing high labour and time costs, leading to delays in repairs due to slow detection of damage. The high costs can be significantly reduced by using latest advancements in computer vision and artificial intelligence (AI) to automate the detection of damage for a timely and low-cost inspection of the vast road infrastructure.

In our previous work "AI-Driven Road Maintenance Inspection" (Mukherjee et al. 2021), we proposed a method to automate the inspection of road infrastructure using AI models trained on commercially viable datasets. However, we identify key areas that are critical for real-world deployment scenarios. First, real-world data tends to be out-of-distribution with respect to the publicly available and commercially viable datasets that are used to create AI models and creating a new dataset every time adds to the cost of the inspection. Second, a bounding box detection indicates the location and presence of the damage but does not provide any quantification for the extent of the damage. This quantification is vital in marking urgency of repairing the damage, the lack of which can lead to issues. For example, minor road damage which may not require immediate repair efforts might get prioritized and consume is proportionate to the extent of damage. Meanwhile, a severe damage elsewhere would get delayed repairs and potentially jeopardize the safety of road users. Both of these issues add manual efforts that can be dispensed with. Hence, there is a need for an optimal, efficient, and automated solution to detecting and quantifying road-damages.

To this end, we present solutions for both the lack of labelled out-of-distribution data, as well as the instance level quantification of the damage. Our main contributions are summarized as follows:

---

[1] Correspondence Email: haris.iqbal@navinfo.eu

- A pipeline that integrates various state-of-the-art AI techniques from semantic segmentation and few-shot learning domains to reduce the time and effort needed to create labelled data to increase robustness of machine learning models in real-life scenarios (usually unseen/out-of-distribution data).
- A method to map pixel-level information that is inconsistent across frames into more tangible and consistent units of measurement using state-of-the-art AI techniques in *Monocular Depth Estimation* and 3D computer vision techniques. We call this approach "*Scaled Camera View Transformation.*"
- We propose a customizable risk evaluation formula, "Instance Quantification of Road Damage", that assigns a risk score to each instance of road damage automatically. This helps in prioritization for resource allocation in road management and repair. The risk-evaluation formula can be customized to meet the user requirement.

## BACKGROUND

In recent years, AI-powered computer vision models have consistently outperformed classical vision methods for scene understanding. Publicly available datasets such as Cityscapes (Cordts et al. 2016) and BDD100K (Yu et al. 2018) are just two examples of driving datasets that can help us create AI models that outperform state-of-the-art methods to solve vision tasks such as object detection and semantic image segmentation.

Despite the remarkable performance of AI models on standard road datasets, the model performance tends to suffer on out-of-distribution (OOD) data i.e., images that are vastly different from the images contained in the dataset used to create the model. One solution to this approach is to create a new labelled data for training models for the same task. However, labelling a new dataset is very expensive and time-consuming.

In addition to the model performance, the real-world road scenarios often require information that is not readily provided by AI models and therefore, needs post-processing. For example, a detection AI model would detect road damage in an image but cannot directly provide output in real-world coordinates. Hence, the estimation of the size of a damage and its associated risk factor cannot be done using AI detection model.

In this paper, we propose a pipeline that uses few-shot learning to deal with the performance reduction on OOD data. Our pipeline can generalize well on a vast range of driving scenarios while minimizing the effort required to annotate large amounts of training data to improve or create new models. Thus, reducing both the overall time and cost investment required to create AI models that are viable for real-life scenarios.

We additionally propose a risk estimation module for road damage that works on detection AI model output. It uses image processing and an AI technique called monocular depth estimation to convert the detection model output to a usable format. Not only it segments the exact damaged area from a detection bounding box, but also scale it from pixel-to-meters, thus providing a scaled estimate of the damage. The risk calculated this way can help properly prioritize damages and allocate repairs efficiently.

## APPROACH

The approach adapts a Road Damage Detection (RDD) model to new OOD data automatically and provide a reliable risk estimate for road damages from the RDD model output. Given the limited labelled road damage data, the RDD model is first trained on it and used in automated labelling pipeline to label the unlabelled data. The unlabelled data consists of images from both diverse statistics and the OOD we want to adapt to. Once the unlabelled data is labelled, the RDD model is trained on the union of both the dataset. This robust RDD model is then used for inferencing on the test data to detect road damages. Here, the road damage detections are in the form of bounding boxes or crops. These crops are then segmented to get exact pixels of damage. Meanwhile, a scaled transformation is also estimated to map these pixel damages to meters to get each damage instance quantified, and assigned a risk factor. The overall pipeline of our approach is shown in Figure 1.

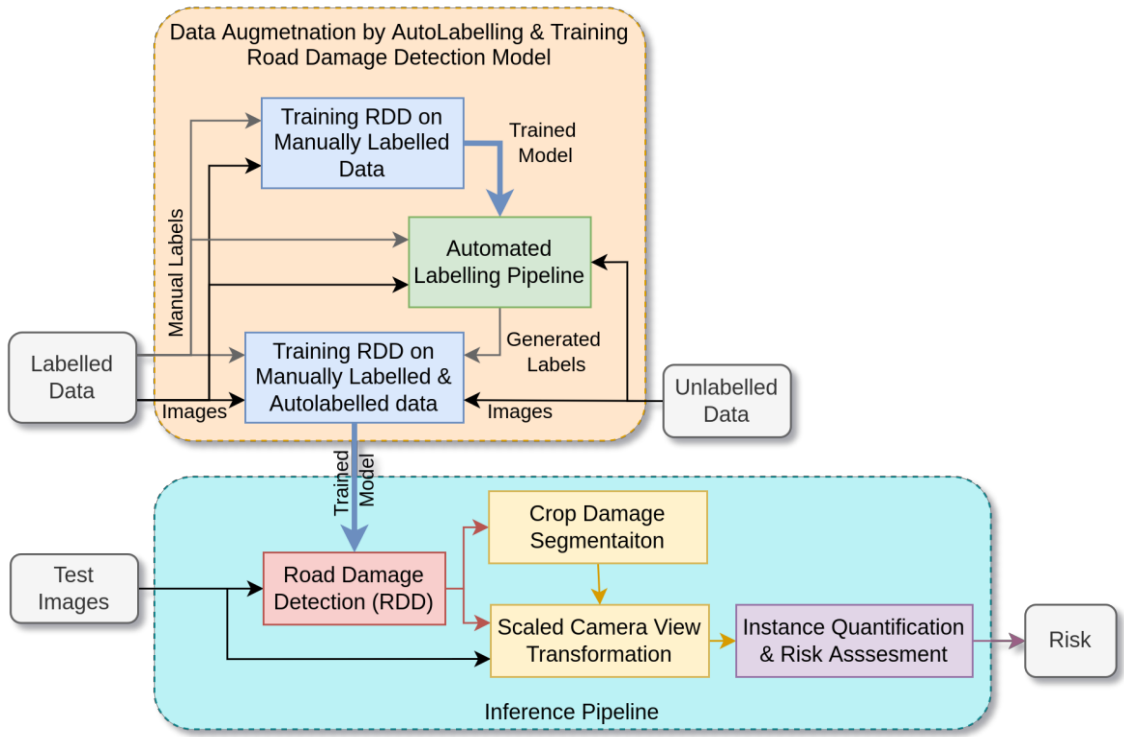

*Figure 1: A schematic overview of the whole road damage detection pipeline for real-world OOD scenario. The pipeline generates robust road damage detection (RDD) model using data generated by our auto labelling pipeline. It also processes the RDD model output to estimate the scale and hence, the risk of a particular damage.*

### Dataset Preparation

***Labelled Data:*** We explored publicly available and commercially viable datasets for road damages and found the Road Damage Dataset (Arya et al. 2020) to be most useful. It consists of images of road damages taken from India, Czech Republic, and Japan, by vehicle-mounted smartphones for low-cost development of road damage detection. The dataset captures four types of road damages considered - longitudinal cracks, transverse cracks, alligator cracks, and potholes. The variety of road structures and corresponding damages provided by this dataset covers most of the damages expected on the road. This would imply that a model trained on this dataset should perform well in almost every scenario and hardly any new data would be OOD. While this may be true for the semantic variation of damages, the statistical variations could still prove to be out-of-distribution, thereby reducing accuracy. In practice, the shift in distribution occurs because of many factors like the different camera setups, collection time, etc by road maintenance agencies, thereby reducing the model performance on new data.

***Unlabelled Data***: For the unlabelled data, we selected images from Mapillary-Vistas dataset (Neuhold 2017), and an in-house real-world task specific video. Mapillary-Vistas provides driving scenario images from all over the world from multiple cameras and therefore has high statistical variation. The in-house video, meanwhile, is selected from a road damage inspection task. This video is about 10mins long and has road damages of all the types considered in this paper i.e., longitudinal cracks, transverse cracks, alligator cracks, and potholes. This video is considered the OOD data we want to our model to adapt to. So, these two datasets, if labelled, are expected to cover both breadth and depth of required statistics, i.e., both general-purpose as well as task-specific information.

### Auto-labelling for Road Damage Detection

To reduce the labour-intensive human involvement in labelling road damage in unlabelled out-of-distribution data, we propose an auto-labelling pipeline that can generate reliable detection labels without human intervention. The pipeline significantly improves upon the naïve approach of sourcing labels by only performing inference on the unlabelled data using the road damage detection model that was trained on the existing labelled data. The increased reliability and retrieval rate of our pipeline make the generated labels suitable as additional training data for the road damage detection task. The schematic overview of the pipeline is shown in Figure 2.

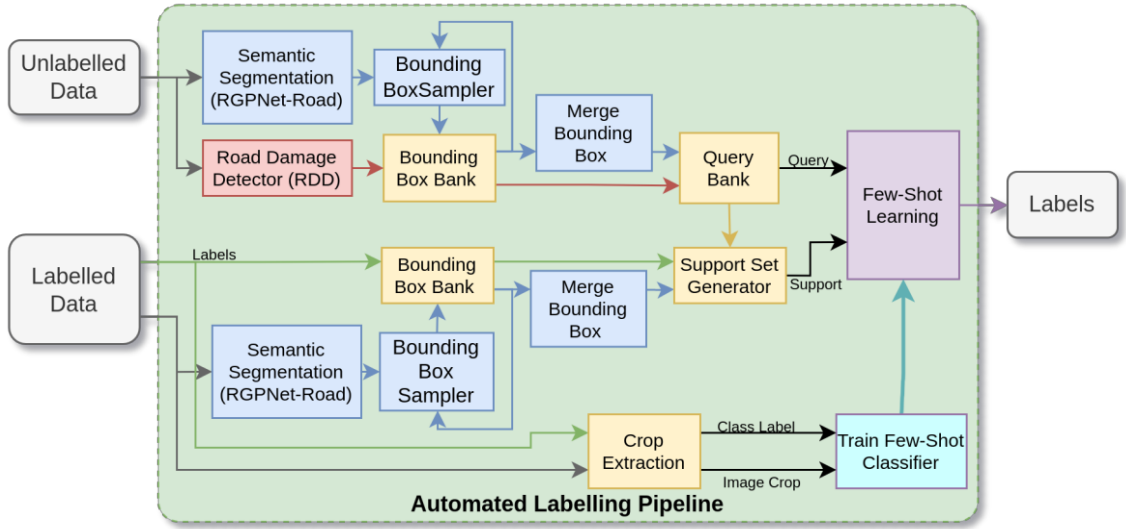

*Figure 2: A schematic overview of our auto-labelling pipeline. The pipeline generates reliable labels of road damage instances in unlabelled image data by sourcing candidate bounding boxes from the unlabelled data and verifying them using few-shot learning. The pipeline only relies on a small labelled dataset for the task at hand which does not have to be sourced from the same data distribution as the unlabelled data.*

The pipeline works by first proposing bounding-box candidates in every image of the unlabelled data using the pre-trained road damage detection model as well as our *Bounding Box Sampler* (*BBS*) module. Thereafter, the pipeline uses our *Few-Shot Classification* (*FSC*) module to assign each of these candidate bounding boxes to the correct road damage type if it is a positive sample or filter it out if it is a negative sample.

The purpose of the BBS module is to generate candidate road damage detections that can potentially serve as detection labels. The sampling of the proposal bounding boxes works by first using a semantic segmentation model to segment the road region in the image. Specifically, we use our in-house state-of-the-art semantic segmentation model, RGPNet (Arani et al. 2021), trained on mapillary-vistas for this purpose. Then, we additionally obtain a mask of the road region along with all the bounding boxes we already have sampled by now. Next, a random pixel is sampled from within the road region, and a random bounding box is placed around it. The size and aspect ratio of the bounding box are sampled to be comparable to the ones in the labelled dataset. This 'without-replacement' Bounding Box Sampler can be employed to get samples from regions where we do not have any prior knowledge like manual labels.

We also have another pipeline proposed for labelling (Brouns et al. 2022), which samples the bounding box differently by using a proposal generating tracked object detection module and therefore, is suitable for tasks related to samples without a background providing prior knowledge. Here, we have prior information of where the road is and therefore, we designed a pipeline around extracting this information.

The proposed bounding boxes are then sent to the *FSC* module which determines whether any road damage is present in the bounding box or not, and if it is, additionally classifies the type of road damage. To make this classification process generalize well to OOD data, we use a *few-shot learning* technique, called *CrossTransformers* (Doersch et al. 2020). Unlike the usual method of training where the model is trained end-to-end on training data and validated on labelled validation data, few-shot learning involves learning from a given small subset of the labelled data as reference and making predictions based on those references. A pretrained feature extractor, trained on road damage dataset in our case, is used to aid the few-shot learning. The subset of labelled images that is given to the model as a reference is called a *support set* while the unlabelled images to be processed are called the *query set*. The few-shot learning method matches spatial feature correspondences between the query and the support set to find the nearest neighbour in feature space, which is then predicted as the label for the query. This approach mitigates the variation in statistics problem discussed earlier, if the support set is chosen appropriately.

The support set should comprise of both positive and negative examples, corresponding to crops of damaged and undamaged road, respectively, and should cover all road damage types. Additionally, the support set should have the same underlying statistics as the query set, to ensure reliable few-shot prediction. For example, if the query set is sampled from an image of a dirt road, our support set should also contain crops extracted from dirt road images. We achieve this by first sourcing a pool of positive and negative examples from the labelled data. We use the ground-truth samples from the labelled data as the positive examples directly, while the negative examples are extracted from the labelled data using the *BBS*. Then, we use a small module to sample the support set from this pool of positive and negative examples, given a query. This module utilizes the colour and gradient histograms of the image to find the nearest positive and negative examples in the labelled data to generate a support set for a particular query. These support and query sets are then fed to few-shot learning method which assigns one of the labels from the support set to each queried image crop.

**Instance Quantification of Road Damage**

In addition to traditional metrics like precision and recall, there is also a need to quantify road damage instances to identify locations of urgency for repairs and other safety measures. Damage and urgency quantification in road maintenance is desirable to allocate limited resources such as manpower and equipment efficiently. As a result, lack of proper prioritization and targeted maintenance can lead to increased costs while increasing traffic risks. For instance, mild damage can be identified for future inspection and need not be repaired immediately. However, a deep pothole can be a safety risk and should be immediately repaired. Therefore, a proper prioritization for maintenance is necessary to allocate repair efforts efficiently, which in turn requires quantification of the road-damage.

To make automatic prioritization of road damages possible, we define the instance quantification metrics for the corresponding road damages in Table 1. These are later used to score each road damage instance allowing the user to automate road maintenance efficiently.

| Damage Type | Quantification Metric |
|---|---|
| Pothole | Area (m$^2$) |
| | Depth: shallow vs deep |
| Longitudinal & Transverse Crack | Length(m) |
| Alligator Cracks | Area (m$^2$) |
| | Crack density (% Area) |

*Table 1: The instance quantification metrics.*

**Risk Factor Assessment**

Depending on the use case, it makes sense to translate these quantities into a perception of risk or repair urgency. For this, we propose an extensible way that can be easily modified in case of different importance criteria:

***Pothole:*** In terms of area, we assign the risk factor linearly. However, for deep potholes their urgency grows rapidly as soon as their area crosses a certain threshold. Hence,

$$Risk_{Pothole}(deep) = e^{Area}$$
$$Risk_{Pothole}(shallow) = Area$$

***Longitudinal & Transverse (Linear) Cracks:*** In a linear crack, we assign a linear factor of risk to the length of the crack.

$$Risk_{LinearCrack} = Length$$

***Alligator Cracks:*** For the alligator crack, there are three factors to consider. One is the area of road damaged, and the other two are the total crack length and pothole like damage blobs. In our calculations, we estimate the last two together as crack density. It is proportional to the length of the total crack in the web as well as area of the blobs. Thus, we propose:

$$Risk_{AlligatorCrack} = Area \times CrackDensity$$

Suitable thresholds can be used to translate Risk into discrete quantities like *Low, Medium and High Risks.*

**Detection Post-processing for Instance Quantification Metrics**

Our road damage detection model gives output in the form of a bounding box with the damage type. A bounding box around the damage is not precise enough to get the size of the damage. This bounding box only corresponds to a crop region in the input image where we have the damage. Moreover, the size estimated from this bounding box would be based on pixels, which tend to be inconsistent across and even within frames, as opposed to metric units (such as meter). Therefore, the damage region is segmented using our crop damage segmentation module from the bounding-box crop as shown in Figure 3. Once the pothole or a crack is segmented, it is mapped from 2D image coordinates to the consistent 3D road coordinates because the road is normally at a perspective angle with respect to camera. The perspective distortion thus introduced makes a nearby pothole appear larger than a farther pothole of the same size within a 2D image frame. However, this problem does not exist in 3D road coordinates. Hence, we need two modules: a *damage segmentor* and an *image-to-road coordinate transformation*, to convert the output from road damage detector into a form that can be used to calculate instance quantification metrics.

*Damage Segmentation*

From a given crop of image, the damaged road segment (crack or pothole) is separated from the undamaged background road region. For this, classical 2D image processing techniques as described in Figure 3 are used. The Image processing techniques suffer from noise and presence of road markings in the crop. For this, our in-house semantic segmentation model – RGPNet is used to separate road markings and processed with different hyperparameters than the ones used to process road region. Additionally, to make the damage segmentation robust to road texture and pixel noise, the crop is also denoised using an edge-preserving filter- Bilateral smoothing.

Once the crop is preprocessed and separated into road marking and background road region, edges are detected on both of these regions with respected hyper-parameters. Normally, edge detection is done with derivative filters of first or higher order. Here, thresholding is considered a special case of edge detection where the order of derivative is zero. It helps not only in segmenting potholes, but also thick cracks in the road and therefore, has a notion of edge detection associated with it. For the zero-order edges, Otsu's thresholding is used because it calculates threshold automatically by minimizing intra-class variance and maximizing the inter-class variance. For the first-order edges, Sobel edge filter provides an compute efficient and decent estimation of edges. Both of these edges are then processed with respective morphological filters to refine the output and filter out the noise.

The edge estimates are then fused using order aggregation and then mask collation. Since a pothole is better segmented using threshold, while a crack is better segmented from edges, the aggregation of order should be dependent on the type of damage being segmented. Therefore, potholes are primarily segmented with zero-order edges while the linear cracks, both longitudinal and transverse, are segmented using first-order edges. However, the alligator cracks consist of a network of linear cracks with potholes like damage in some regions. Therefore, we allocate a higher weight to the zero-order edges while the first-order edges are still considered with lower weight. Finally, using the road marking mask, damage segmentations from both the road marking and non-marking regions are collated into one segmentation of the whole crop as shown in Figure 4 and 5.

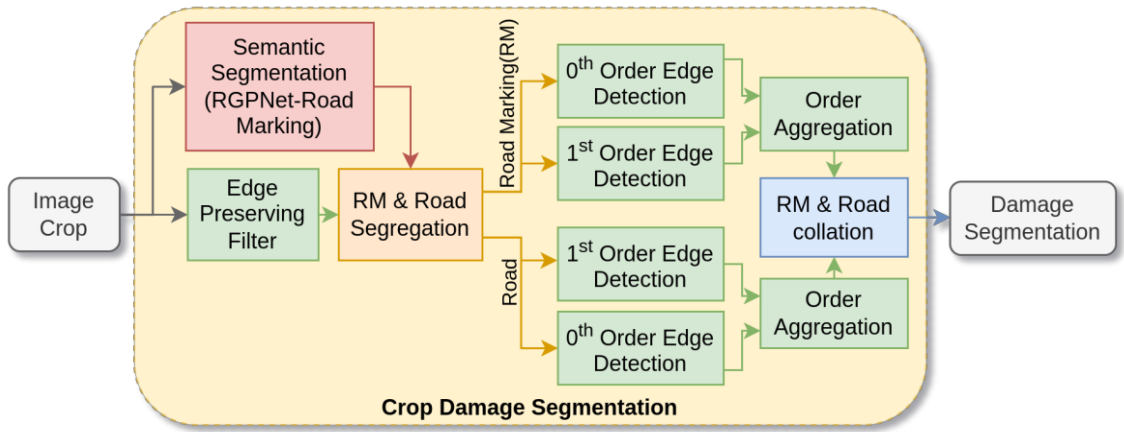

*Figure 3: Schematic overview of our Road Damage Segmentation from image crops.*

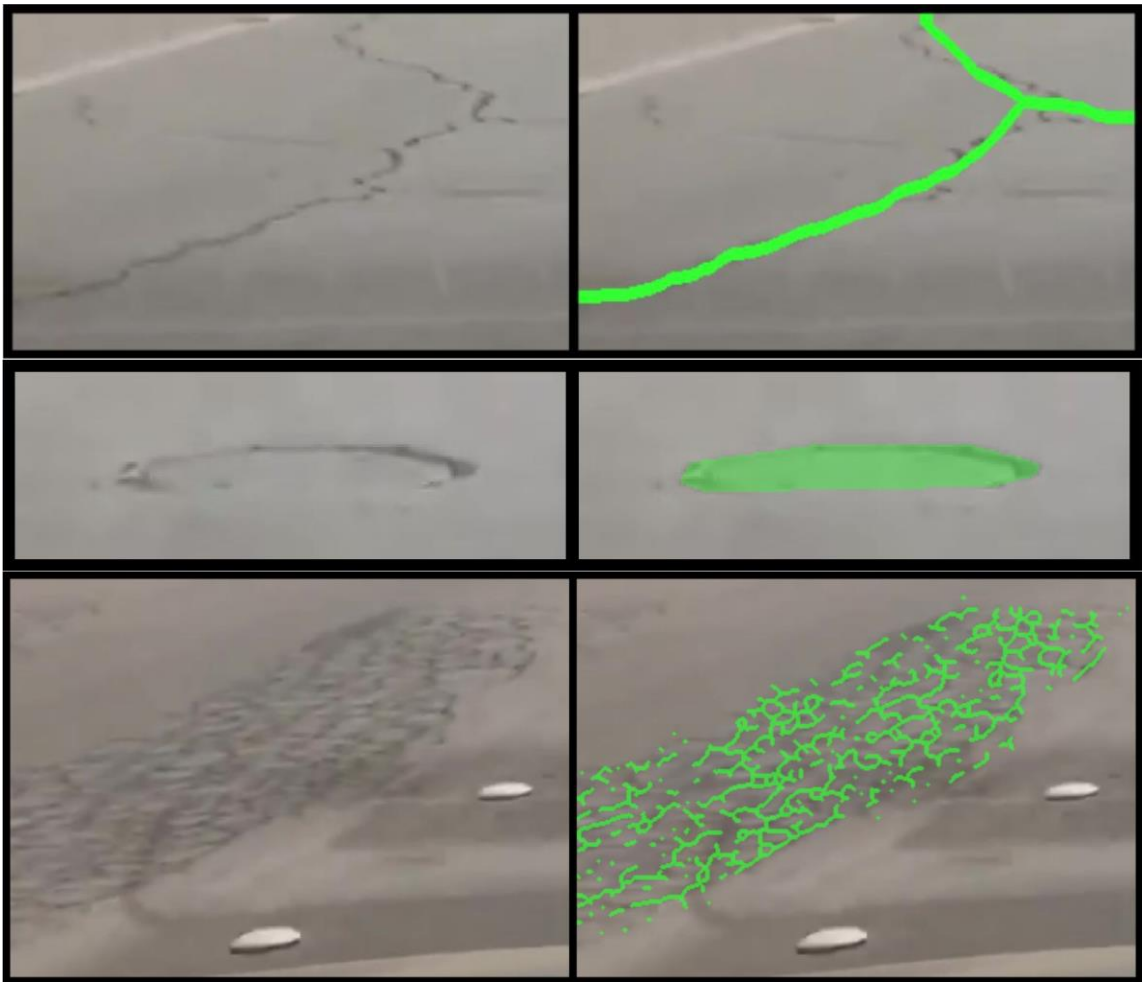

*Figure 4: (left) input crop passed to the Crop Damage Segmentation module, (right) Segmented Damage (Linear crack, pothole and alligator crack from top-to-bottom) shown in Green overlayed on the input crop.*

Finally, the segmentation maps need to be combined with 3D knowledge to quantify the road damage. To do so robustly and consistently, we need to remove perspective distortion from a given road crop in an image. This is done by projecting the viewing angle of the camera to the scaled bird's-eye-view as shown in Figure 5.

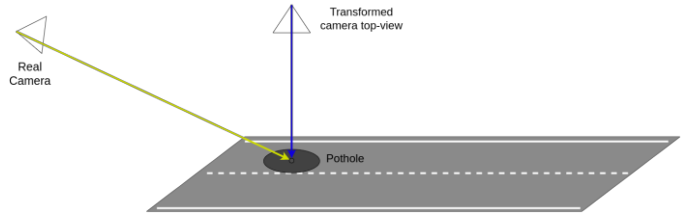

*Figure 5: A visualization of the aim of the Scaled Camera View Transformation. Input videos are taken from an angle shown by the "Real Camera" in the figure whereas, we want to have top-view for our calculations.*

*Scaled Camera View Transformation*

There are two aspects to this problem. First, to rotate and translate the viewing angle. Second, to estimate the meter to pixel scale. Both of these can be estimated using unsupervised monocular depth estimation and the known camera height. If camera height is not known, same can be achieved using high quality geotagged videos of similar scenarios to train a scale-consistent unsupervised monocular depth estimation model (Chawla et al. 2021). In that case, the depth output from the model will already be scaled. In Figure 6, we show the whole pipeline of transforming the viewing angle to top-view.

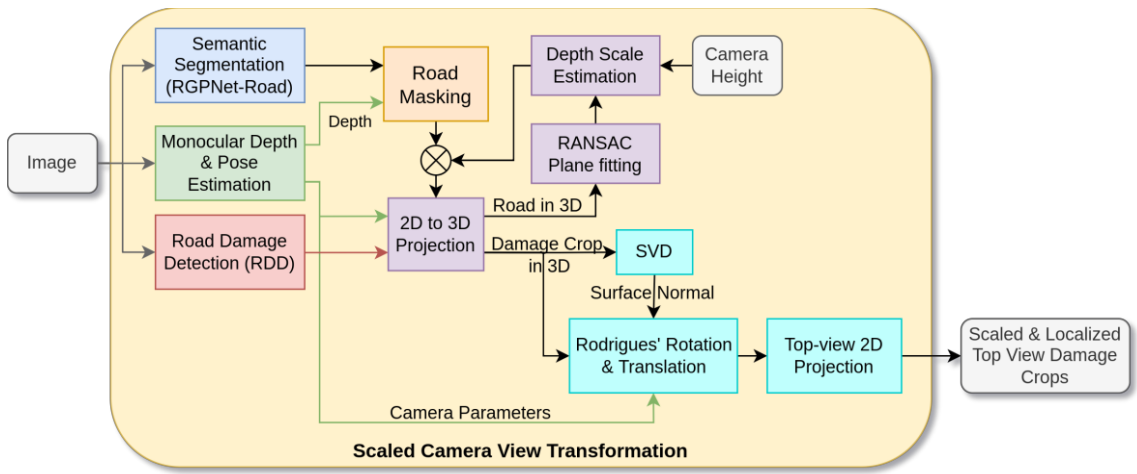

*Figure 6: Schematic overview of our Scaled Camera View Transformation of image crops. Additional components which are part of the whole pipeline like Road Damage Detector and Semantic Segmentation are also shown.*

The algorithm (assuming scaled depth estimates are not available) is described as follows:

- An image is passed to the *Semantic Segmentation model* to get *road mask* and a *Monocular Depth Estimator* to get depth map. The *depth map* is then masked by the *road mask* to get depth mask of just the road pixels.
- The *Monocular Depth Estimator* MT-SfMLearner (Varma et al. 2022) also predicts the camera intrinsics, which is used to transform the road pixels from a 2D space to 3D space using the depth information and the perspective projection equation.

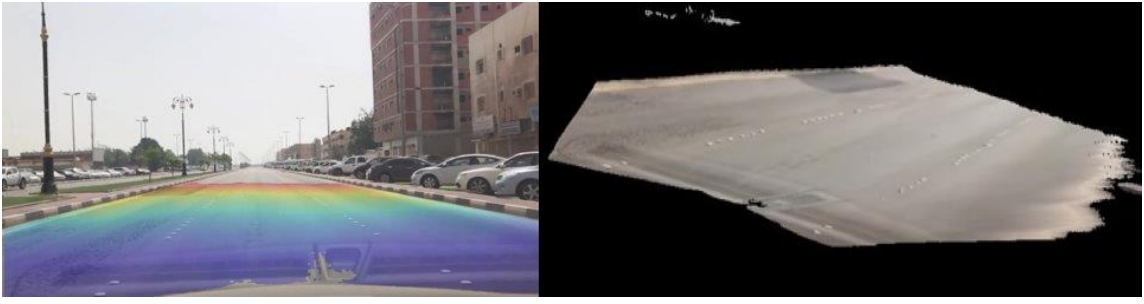

*Figure 7: (left) An example input image with depth map for just the road pixels overlaid. (right) Corresponding 3D point-cloud of the road.*

- Random Sample Consensus (RANSAC) algorithm is used for fitting a plane on 3D point cloud of road pixels obtained in the previous step. The equation of this plane is:

$$ax + by + cz = d$$

- The known height $h$ (in meters) of the camera i.e., the distance between road plane and camera center, is used to estimate the scale of the plane (McCraith et al. 2020). Let origin (0,0,0) be at the camera center, (0, -h, 0) be the point on the road directly below the camera, and (0, -k*h, 0) be the point directly below the camera on the unscaled plane. Then, the scale k is calculated as:

$$a(0) + b(-kh) + c(0) = d \Rightarrow k = \left|\frac{d}{bh}\right|$$

| Scaled Plane Parameters | a | b | c | d |
|---|---|---|---|---|
| Mean | 0.066 | 0.914 | 0.4 | 1.188 |
| Standard Deviation | 0.046 | 0.013 | 0.022 | 0.017 |

*Table 3: As expected the value of b is close to 1 representing the normal vector of the road plane is upwards. Additionally, c=0.4 represents that the camera is slightly inclined towards road and is not completely horizontal. Camera Height used for scaling is 1.3 m.*

- The scaled depth map (in metres) is used to project each bounding box from the *Road Damage Detector* to the scaled 3D world (called camera world) using camera intrinsics obtained from the *Monocular Depth Estimator*.
- Singular Value Decomposition (SVD) of the 3D point cloud of the cropped road damage is used to find its *Normal Vector*. This is done because while the road is planar, road damages will form local variations on the plane and the normal estimated from road-damage would be more robust.
- Rodrigues' vector rotation formula is used to rotate the camera-axis to align the camera with the normal direction vector of the road-damage crop.
- The camera-center is translated from the origin to lie on the *Normal Vector* passing through the centroid of the road-damage crop. Finally, an orthographic projection is used to project the scaled 3D point cloud to a 2D top-view image as shown in *Figure 8*. Orthographic projection retains the original sizes of objects while reducing the impact of perspective distortions as discussed earlier (See *Figure 5*).

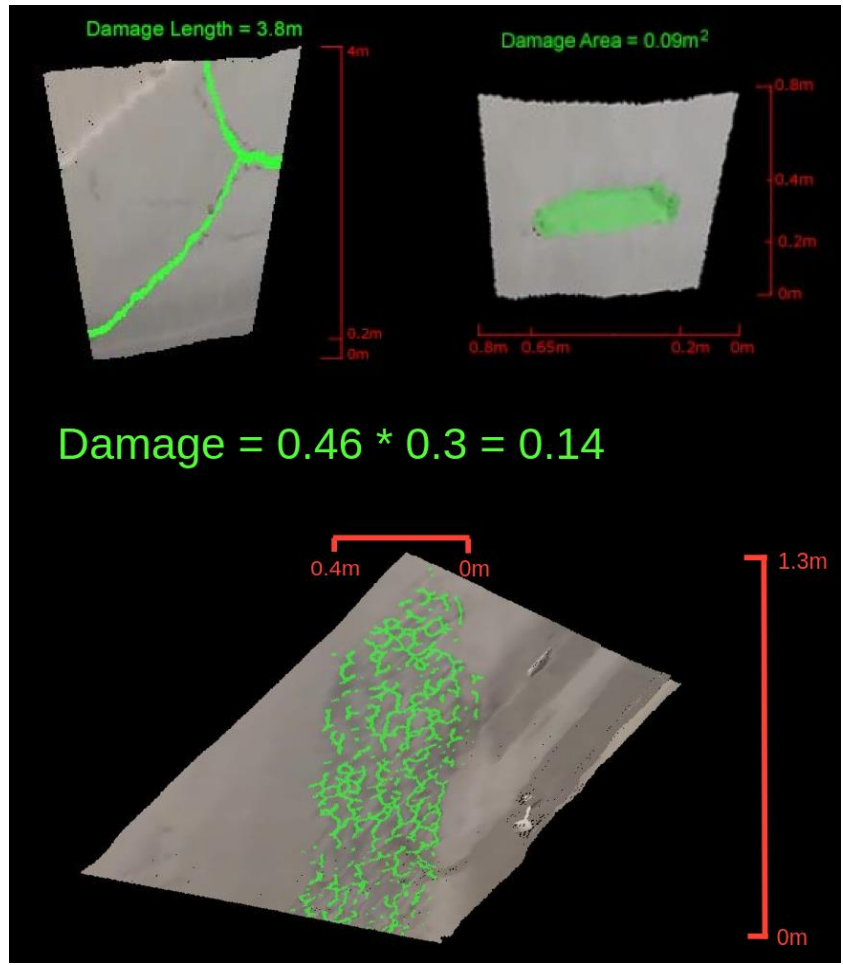

*Figure 8: Output from our Crop Damage Segmentation module overlayed on the Scaled top-view transformation of the damage crops. The length of the crack can now be easily calculated. The inputs are shown in Figure 4. All of these are considered medium risk damages. It should be noted that a rectangular box in perspective view would be distorted to quadrilateral in top-view as shown.*

**MAP VISUALIZATIONS**

For quick inspection, we visualize the results on the map. First, we extract the GPS information from the videos and then, interpolate it to gain GPS coordinates for each frame of the test video. The road damages are marked for the specific frames and matched with their corresponding GPS coordinates. In the end, the locations of each road damage are pinned along with an indicator of risk factor as shown in Figure 9.

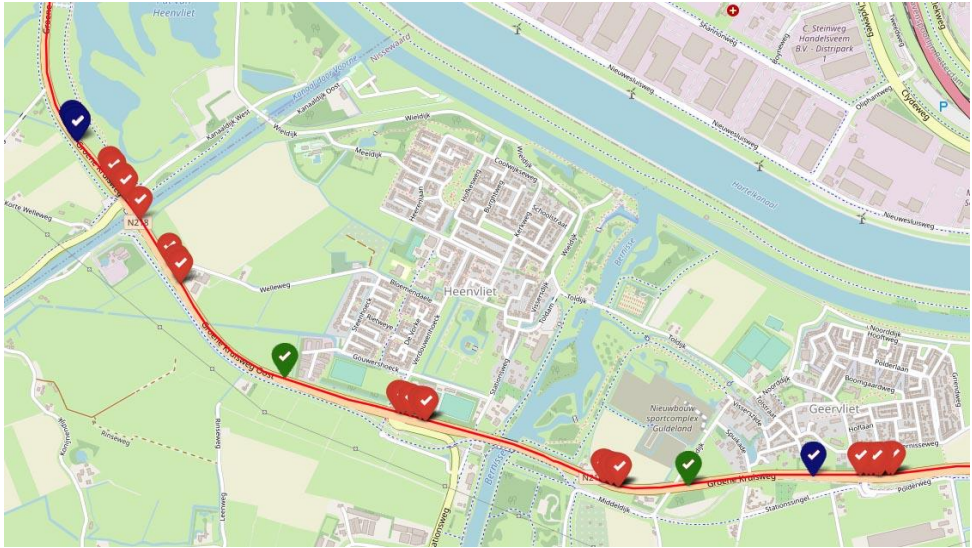

*Figure 9: Example visualization of precise geographical locations of the damages along with their risk level. Green, blue, and red pins show low, medium, and high-risk damage, respectively.*

## CONCLUSION

We presented two approaches to reduce the labor and time cost of road maintenance inspection using AI techniques. The first one alleviates the labelled data requirement for road damage detection by automating the data generation. The second one improves the maintenance deployment efficiency by automated quantification of the extent of a damage instance and assigning a risk factor to it.

In the future, other techniques can be explored to improve both the detection accuracy and the quantification of instance damage. The detection accuracy can be increased by improving the automated labelling pipeline. One direction could be to include human-in-the-loop to improve the quality of labels generated in a semi-automated way. For quantification of instance damage, we can explore stereo camera, which is still much cheaper than a LiDAR, to increase precision and accuracy of 3D reconstruction and consequently localization of damage segment. This can increase accuracy of road damage detection while not increasing the cost significantly.